\documentclass[acmsmall,screen]{acmart}

\usepackage{bm}
\usepackage{subfig}
\usepackage[ruled,vlined,linesnumbered]{algorithm2e}
\usepackage{multirow}

\setcopyright{acmcopyright}
\copyrightyear{2019}
\acmYear{2019}
\acmDOI{10.1145/1122445.1122456}

\acmJournal{TIST}
\acmVolume{37}
\acmNumber{4}
\acmArticle{111}
\acmMonth{8}

\begin{document}

\title{CAGNN: Cluster-Aware Graph Neural Networks for Unsupervised Graph Representation Learning}

\author{Yanqiao Zhu}
\authornote{Both authors contributed equally to this research.}
\email{yanqiao.zhu@cripac.ia.ac.cn}
\orcid{0000-0003-2205-5304}
\affiliation{
	\institution{Center for Research on Intelligent Perception and Computing, Institute of Automation, Chinese Academy of Sciences}
	\city{Beijing}
	\state{China}
}
\affiliation{
	\institution{School of Artificial Intelligence, University of Chinese Academy of Sciences}
	\city{Beijing}
	\state{China}
}

\author{Yichen Xu}
\authornotemark[1]
%\authornote{This work is done during his internship at CRIPAC, CASIA.}
\email{linyxus@bupt.edu.cn}
\affiliation{
	\institution{School of Computer Science, Beijing University of Posts and Telecommunications}
	\city{Beijing}
	\state{China}
}

\author{Feng Yu}
\email{yf271406@alibaba-inc.com}
\affiliation{
	\institution{Alibaba Group}
	\city{Beijing}
	\state{China}
}

\author{Shu Wu}
\email{shu.wu@nlpr.ia.ac.cn}
\author{Liang Wang}
\email{wangliang@nlpr.ia.ac.cn}
\affiliation{
	\institution{Center for Research on Intelligent Perception and Computing, Institute of Automation, Chinese Academy of Sciences}
	\city{Beijing}
	\state{China}
}
\affiliation{
	\institution{School of Artificial Intelligence, University of Chinese Academy of Sciences}
	\city{Beijing}
	\state{China}
}

\renewcommand{\shortauthors}{Zhu et al.}

\begin{abstract}
Unsupervised graph representation learning aims to learn low-dimensional node embeddings without supervision while preserving graph topological structures and node attributive features. Previous graph neural networks (GNN) require a large number of labeled nodes, which may not be accessible in real-world graph data. In this paper, we present a novel cluster-aware graph neural network (CAGNN) model for unsupervised graph representation learning using self-supervised techniques. In CAGNN, we perform clustering on the node embeddings and update the model parameters by predicting the cluster assignments. Moreover, we observe that graphs often contain inter-class edges, which mislead the GNN model to aggregate noisy information from neighborhood nodes. We further refine the graph topology by strengthening intra-class edges and reducing node connections between different classes based on cluster labels, which better preserves cluster structures in the embedding space. We conduct comprehensive experiments on two benchmark tasks using real-world datasets. The results demonstrate the superior performance of the proposed model over existing baseline methods. Notably, our model gains over 7\% improvements in terms of accuracy on node clustering over state-of-the-arts.
\end{abstract}




\maketitle

\section{Introduction}

Unsupervised graph representation learning aims to learn low-dimensional node embeddings without supervision. The learned node embeddings preserve useful topological structures and node attributive features extracted from graphs.
%Extensive work demonstrates that a variety of downstream graph analytical tasks, such as multi-label classification \cite{Tang:2015ew,Grover:2016ex} and community detection \cite{Sun:2019ul,Li:2018wg}, can greatly benefit from these learnt node representations.
Traditional graph representation learning algorithms originate in the skip-gram model for distributed language representation \cite{Mikolov:2013uz}. The pioneering work DeepWalk \cite{Perozzi:2014ib} constructs node sequences by performing random walks over the graph. Then, on top of these sequences, the node embeddings can be learned using the skip-gram model. Following this line of development, various network embedding methods have been proposed, such as node2vec \cite{Grover:2016ex} and LINE \cite{Tang:2015ew}.

Recently, the graph neural network (GNN), a generalized form of convolutional networks in the graph domain, has attracted a lot of attention. Compared with conventional graph embedding methods, GNN shows superior expressive power and has achieved promising performance in many tasks \cite{Kipf:2016tc,Velickovic:2019tu,Chen:2018vh,Xu:2019ty}. However, most existing GNN models are established on a semi-supervised setting \cite{Kipf:2016tc,Velickovic:2018we}. Training an accurate GNN model requires a number of high-quality node labels, which might not be accessible. Then, a natural question is that \emph{can we leverage the expressive power of GNN models and produce node embeddings in an unsupervised manner}?
In the real world, graphs can be derived from business data in quantity, which can help facilitate analytical tasks and provide valuable insights for business. Take an e-commerce website for example, structured data derived from every-day user purchases along with their relationship with items is produced in a million scale. Given the purchase data, we can better classify users and items if we further leverage the derived interaction graphs. Since obtaining labels is a labor-intensive and time-consuming process, if we can efficiently train a GNN model in an unsupervised manner, it would be greatly beneficial to facilitate downstream analytical tasks.

There has been a surge of research interest in unsupervised visual representation learning to avoid expensive data annotations. As a subfield of unsupervised learning, self-supervised methods have achieved great success \cite{Pathak:2016gb,Noroozi:2016hd,Noroozi:2018bf}.
Among them, a series of work performs clustering on embeddings and regards the clustering assignments as the pseudo-labels to replace human annotations. Then, the classification objective can be used to train the model.
% Specifically, they perform clustering on the embeddings and regard the clustering assignments as their pseudo-labels, since clustering requires little domain knowledge and {\color{blue} cluster labels provide useful structure information}.
The work DeepCluster \cite{Caron:2018ba} uses \(k\)Means to compute pseudo-labels from raw image data, enabling large-scale visual representation learning in an unsupervised manner. Following this work, \citet{Asano:2020ww} further propose a self-labeling scheme to regularize the cluster size and avoid degenerate solutions, which has become the state-of-the-art method on computer vision benchmarks.

Although there is a proliferation of studies in self-supervised visual representation learning, little attention is paid to graph representation learning using a self-supervised manner.
\citet{Hu:2020ws} propose a pre-training GNN model that is trained using several network measures such as betweenness and closeness. However, these statistical measures require domain knowledge and are sensitive to noise in graphs. On the contrary, as a natural characteristic of graph data, clusters group vertices that share similar functionalities in a graph and thus can be used as a good supervisory signal for training the GNN model.
Moreover, we observe that graphs often contain noisy edges, which connect nodes belonging to different classes. Such edges may mislead GNN training and further confine the model from learning useful class information. In a graph with many inter-class edges, when performing graph convolution through neighborhood aggregation, i.e. taking the average over neighbor nodes, the resulting node embeddings tend to be indistinguishable from different classes. Thus, we argue that a key to improving the quality of embeddings is to alleviate the impact of potentially noisy edges and strengthen edges between nodes of the same class, which will help preserve the cluster structures and obtain better-separated node embeddings. In summary, considering the existence of noisy inter-class edges, it is crucial to mitigate the impact of these edges during training, which is usually neglected by previous work that merely leverages network measures as self-supervision.
% It is more desirable to adaptively capture class information in a graph while simultaneously perform graph denoising.

% Consider the example illustrated in Figure \ref{fig:strengthening}. For the central node, with the progress of neighborhood aggregation (that is taking average over neighbor nodes), the resulting node embeddings are indistinguishable from different classes. If we can isolate neighbor nodes of different class and strengthen edges between nodes of the same class, we will obtain better-separated embeddings which preserve the cluster structures.
%\citet{Sun:2020wx} propose a multi-stage self-learning model, which incrementally generates labels for unlabeled data from labeled nodes belonging to the same cluster. Their work still requires labeled nodes and thus falls in semi-supervised learning.
%As the graph representation learning paradigm resembles that of visual representation learning, we argue that the success of unsupervised visual representation learning models can be duplicated in deep graph learning domains. Since cluster structures are fundamental and informative in graphs \cite{Tian:2014wl}, it is promising to leverage clusters as pseudo-labels to self-supervise GNN training.

\begin{figure}[t]
	\centering
	\includegraphics[width=\linewidth]{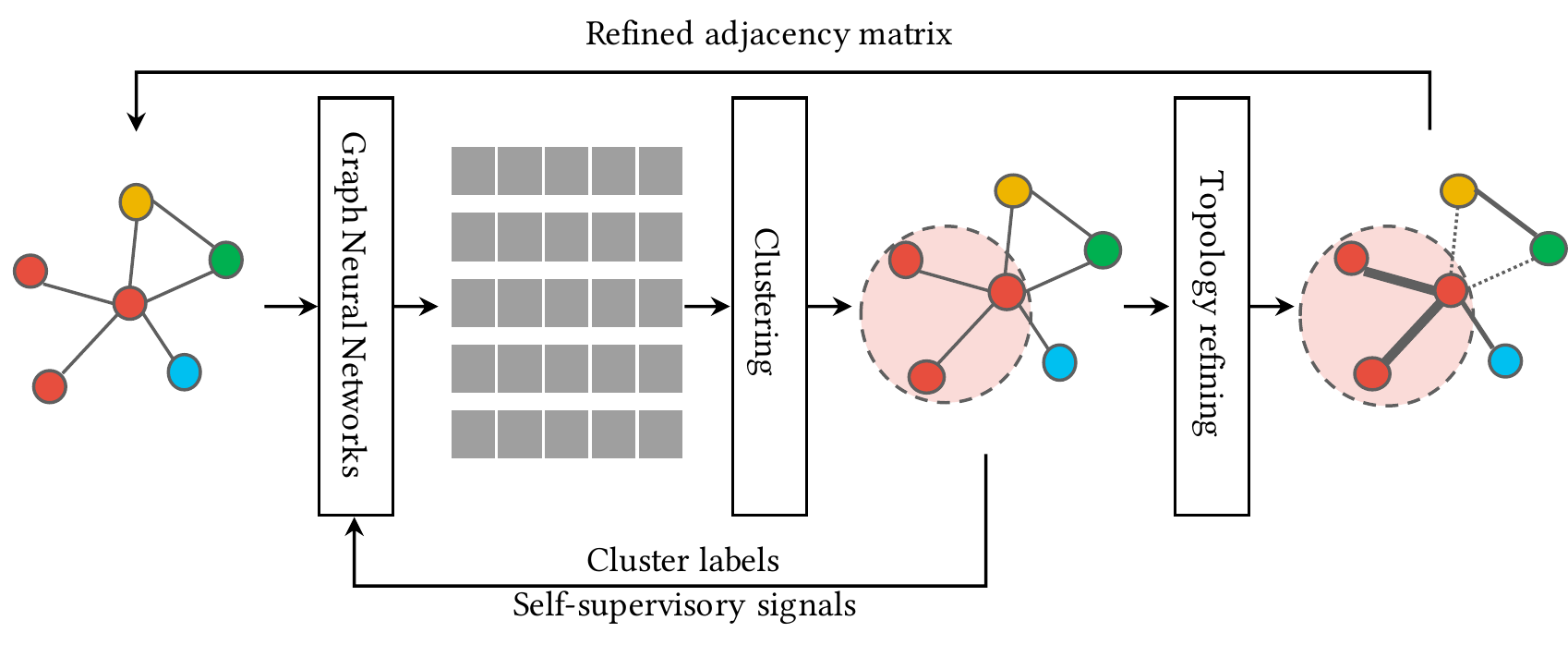}
	\caption{The pipeline of the proposed CAGNN model. The CAGNN model alternates between node representation learning and clustering. We first obtain node embeddings using graph neural networks (GNN). Then, we perform clustering and use the cluster labels as the self-supervisory signals. Following that, we use a novel cluster-aware topology refining mechanism which reduces inter-cluster edges and strengthens intra-class connections to mitigate the impact of noisy edges.}
	\label{fig:model}
\end{figure}

Motivated by the aforementioned observations, we propose a novel cluster-aware graph neural network model for self-supervised graph representation learning in this paper. We term the model CAGNN for brevity. As illustrated in Fig. \ref{fig:model}, our CAGNN model consists of three stages. At the first stage, we perform graph convolutions to obtain node embeddings. Then, the model conducts clustering on the node embeddings and updates the model parameters by predicting the corresponding cluster assignments. To avoid degenerate solutions, we use a balanced cluster strategy, which formulates the cross-entropy minimization as an optimal transport problem. Finally, to alleviate the impact of noisy edges and better preserve cluster structures in the embedding space, we propose a novel graph topology refining scheme based on cluster assignments. The proposed refining process strengthens intra-class edges and weakens potentially noisy edges by isolating neighborhood nodes of different clusters.

The core contribution of this paper is three-fold. Firstly, we propose a novel self-supervised graph neural network for unsupervised graph representation learning, which needs no supervision from labels. Secondly, unlike other GNN models, CAGNN further proposes a topology refining scheme which reduces inter-cluster connections of neighbor nodes to alleviate the impact of noisy edges. Thirdly, extensive experiments conducted on benchmark datasets demonstrate the superiority over existing baseline methods. It is worth mentioning that the proposed method gains over 7\% performance improvement in terms of accuracy on node clustering over state-of-the-arts.

The organization of the remaining of the paper is summarized below. We first review prior arts in relevant domains in Section 2. Then, in Section 3, we introduce our proposed cluster-aware graph neural networks in detail. After that, we present empirical studies in Section 4. Finally, we conclude the paper and point out future research directions in Section 5.

\section{Related Work}

In this section, we firstly review work on representation learning methods on visual data. Following that, we emphasize on representation learning on graphs. Finally, we briefly review representative literature on graph neural networks.

\paragraph{Representation learning on visual data.}
%Abundant visual data such as ImageNet \cite{Deng:2009jn} is nowadays widely available. Aiming to produce general-purpose features without human annotations, unsupervised representation learning attracts a lot of research attention.
%Instead of designing specific rules from raw data
Recent development in deep convolutional neural networks has witnessed a transition from hand-crafted features to end-to-end feature learning. As a promising subfield of unsupervised learning, the self-supervised framework achieves superior performance on many benchmark tasks. Self-supervised learning defines pretext learning tasks that can be constructed from raw data and uses the produced embeddings for downstream machine learning tasks of interest. Many strategies for pretext learning tasks, such as image in-painting \cite{Pathak:2016gb}, jigsaw puzzles \cite{Noroozi:2016hd,Noroozi:2018bf}, grayscale image colorizing \cite{Zhang:2016fr,Larsson:2017vt}, and geometric transformation recognition \cite{Gidaris:2018wr}, have been proposed recently. However, the methods using the classification objective, which minimizes the cross-entropy loss, still obtain the best performance \cite{Caron:2018ba,Asano:2020ww}. Along this line, many methods focus on how to obtain proper labels for the classification task. For example, DeepCluster \cite{Caron:2018ba} is proposed to iteratively cluster images using \(k\)-Means; the cluster assignments are then fed as supervision to train the convolutional network. Recently, \citet{Asano:2020ww} combine representation learning and clustering and propose a novel self-labeling scheme to balance the size of clusters, which outperforms existing methods.

\paragraph{Representation learning on graphs.}
Representation learning on graphs is far more complex than image data since there is no spatial locality in graphs. Random-walk-based methods, one of the most popular research tracks, sample random walk sequences and learn node embeddings using sequential models. Representative methods in this domain include DeepWalk \cite{Perozzi:2014ib}, which uses random walks to sample a series of node sequences from the input graph. Then, based on these random walks, node embeddings are generated using language models. Following their work, node2vec \cite{Grover:2016ex} is proposed to design a biased random walk procedure, adding flexibility in exploring diverse neighbor nodes. \citet{Ribeiro:2017ji} further emphasize on the structural similarity between distant nodes.

Compared with random-walk-based methods, another line of development focuses on matrix factorization techniques. The first work LINE \cite{Tang:2015ew} explicitly factorizes first- and second-order proximities, instead of combining them by sampling fixed-length random walks. By explicitly incorporating community information, M-NMF \cite{Wang:2017vp} factories the proximity matrix and the community modularity matrix. Recently, \citet{Qiu:2018ez} theoretically unify random walks and matrix factorization techniques into a framework. However, these methods suffer from insufficient representation ability, because they generate embedding for each node independently and no parameters are shared between nodes \cite{Hamilton:2017wa}.

\paragraph{Graph neural networks.}
As a generalization of convolutional operations on graphs, in recent years there have been a surge of research interests in graph neural networks. Motivated by graph spectral theory, graph convolutional networks (GCN) \cite{Kipf:2016tc} generalize convolutional operations on graphs, which generate node embeddings by aggregating information from neighborhoods. GraphSAGE \cite{Hamilton:2017tp} samples a fixed size of neighbors instead of leveraging the graph Laplacian matrix to compute node embeddings in an inductive manner. Graph attention networks (GAT) \cite{Velickovic:2018we} further introduces an attentional mechanism to learn edge weights. Recently, SGC \cite{Wu:2019vz} is proposed to reduce the computational complexity by removing the non-linearity of graph convolutional networks. Readers of interest may refer to \cite{Wu:2019tn} for a comprehensive survey on GNN.

Also, there are some methods proposed for combining GNNs and other machine learning methods for different graph analytical tasks. For instance, \citet{Zhang:2019bi} propose to address the graph clustering problem via a plain neighborhood aggregation scheme; their proposed method simply aggregates information from neighborhood nodes, without parameters to learn from data. \citet{Wang:2019dx} propose to use a deep-clustering-based method on node embeddings for graph clustering, where a cluster hardening loss is introduced to emphasize the clusters with high confidence. Moreover, the same deep clustering scheme has been applied to semi-supervised learning with few labels as well \cite{Sun:2020wx}. In this work, the clustering algorithm incrementally generates labels for unlabeled data from labeled nodes belonging to the same cluster.

Considering that there is no ground truth label for the model to rely on, unsupervised representation learning is much more difficult and none of the above methods try to solve the unsupervised representation learning problem on graphs. Our work, on the contrary, aims to learn discriminative features directly from attributes and structures of the graph data without supervision, which is more suitable for large-scale data analytics.

\section{The Proposed Method: CAGNN}

In this section, we first briefly introduce the background of graph representation learning. Then, we describe our proposed CAGNN method in detail. Specifically, our CAGNN method consists of two stages, i.e. self-supervised learning and cluster-aware neighborhood refining. After that, we describe the model training algorithms and present complexity analysis as well.

\subsection{Problem Formulation and Notations}

\begin{table}[b]
 \centering
 \caption{Notations used throughout this paper.}
 \begin{tabular}{cl}
 \toprule
 Notation & Description \\
 \midrule
 \(\mathcal G\)       & the input graph \\
 \(\mathcal V\)       & the set of vertices \\
 \(\mathcal E\) & the set of edges \\
 \(v_i\) & the vertex with index \(i\) \\
 \midrule
 \(\bm A\) & the adjacency matrix of graph \(\mathcal G\) \\
 \(\tilde{\bm A}\) & the adjacency matrix with self-loops added \\
 \(\tilde{\bm D}\) & the degree matrix of \(\tilde{\bm A}\) \\
 \midrule
 \(\bm X\) & the feature matrix \\
 \(\bm x_i\) & the feature of node \(v_i\) \\
 \(\bm H\) & the output embedding matrix \\
 \(\bm h_i\) & the embedding of node \(v_i\) \\
 \(\bm H^{(t)}\) & the output embedding of the \(t\)-th layer \\
 \midrule
 \(y_i\) & the cluster label of \(v_i\) \\
 \(\bm C\) & the cluster-assignment matrix \\
 \(\bm c_i\) & the cluster assignment of \(v_i\) \\
 \(\bm \mu_{y_i}\) & the centroid of \(y_i\) \\
 \midrule
 \(\phi_p(\cdot)\) & graph purity function \\
 \(\tau_a\) & the threshold for adding edge in topology refining module \\
 \(\tau_r\) & the threshold for removing edge in topology refining module \\
 \bottomrule
 \end{tabular}
 \label{tab:notations}
\end{table}

Consider an input graph \(\mathcal{G} = (\mathcal{V}, \mathcal{E})\), where \(\mathcal{V} = \{ v_1, v_2, \dots, v_n \}\) denotes the set of nodes and \(\mathcal{E} \subseteq \mathcal{V} \times \mathcal{V}\) denotes the set of edges. We denote \(\bm{X} \in \mathbb{R}^{n \times m}\) and \(\bm{A} \in \mathbb{R}^{n \times n}\) as the node feature matrix and the adjacency matrix respectively, where \(\bm{A}_{ij} = 1\) if \((v_j, v_i) \in \mathcal{E}\) and \(\bm{A}_{ij} = 0\) otherwise. The goal of graph representation learning is to learn a low-dimensional representation \(\bm{h}_i \in \mathbb{R}^d\) for each node \(v_i \in \mathcal{V}\), where \(d\) is the dimension of node representations and \(d \ll n\). We summarize all notations used throughout this paper in Table \ref{tab:notations} for clarity.

\subsection{Self-supervised Learning on Graphs by Clustering}
Typically, GNN models are trained using the classification objective in a supervised manner. 
In our unsupervised model where no ground-truth labels are given, we generate pseudo-labels to provide self-supervision by iteratively performing clustering on the embeddings.
To be specific, in this self-supervised training phase, CAGNN alternates between optimizing parameters of the model by predicting cluster labels and updating the cluster assignments of nodes.

\paragraph{Graph convolutional networks.}
We use graph convolutional networks (GCN) \cite{Kipf:2016tc} as the base model to generate node embeddings. GCN is a multilayer feedforward network in the graph domain that generates node embeddings by aggregating and transforming information from neighbor nodes. We define \(\bm{H}^{(t)}\) as the output of the \(t\)-th layer. The propagation rule of each layer can be defined as
\begin{equation}
	\label{eq:gcn-rule}
	\bm H^{(t)} = \sigma(\tilde{\bm D}^{- \frac 1 2} \tilde{\bm A} \tilde{\bm D}^{- \frac 1 2} \bm H^{(t-1)} \bm W^{(t)}),
\end{equation}
where \(\tilde{\bm{A}}\) is the normalized adjacency matrix of \(\bm{A}\) with self-loops added, \(\tilde{\bm{D}}\) is the degree matrix for \(\tilde{\bm{A}}\) with entries \(\tilde{\bm{D}}_{ii} = \sum_{j=1}^n \tilde{\bm A}_{ij}\), \(\sigma\) denotes the activation function, e.g., \(\operatorname{ReLU}(\cdot) = \max(0, \cdot)\), and \(\bm W^{(t)}\) is the trainable weight parameter of the \(t\)-th layer. The input of GCN is the feature matrix, i.e. \(\bm{H}^{(0)} = \bm{X}\). We employ an \(l\)-layer GCN to produce node embeddings for nodes, i.e. \(\bm{H} = \bm{H}^{(l)}\).

\paragraph{Clustering on node embeddings.}
In CAGNN, cluster labels are used to provide pseudo-labels for self-supervision. A cluster of nodes in the graph is a group of nodes that are closely correlated in terms of both topology and feature. A variety of clustering methods have been developed, and among them \(k\)-Means is one widely-used algorithm. Assume that the cluster labels of \(v_i \in \mathcal{V}\) is denoted by \(y_i \in \{ 1, 2, \cdots, k\}\), drawn from a space of \(k\) possible clusters. We denote the cluster assignments by \(\bm{C} \in \{0, 1\}^{n \times k}\), where each row represents the cluster assignments of one node using one-hot encoding.
Conventional \(k\)-Means aims to learn the centroids \( \bm{\mu}_1, \bm{\mu}_2, \dots, \bm{\mu}_k \) and cluster assignments \(y_1, \dots, y_n \) by optimizing
\begin{equation}
	\min_{\bm{\mu}_1, \dots, \bm{\mu}_k} \frac{1}{n} \sum_{i=1}^n \min_{\bm{C}} \| \bm{h}_{i} - \bm{\mu}_{y_i} \|.
	\label{prob:kmeans}
\end{equation}

However, some clustering-based self-supervised models that directly adopt \(k\)Means are prone to degenerated solutions \cite{Caron:2018ba}. This can be seen by a fact that when we jointly optimize the embeddings \(\bm{h}_i\), a trivial solution to Problem (\ref{prob:kmeans}) can be obtained by mapping all nodes to the same point in the embedding space and treat them as a single cluster. To alleviate this problem, we relax \(\bm{C}\) to be in \(\mathbb{R}^{n \times k}\) such that each row \(\bm{c}_i\) is a probability distribution of node cluster label, i.e. \(\bm{C1} = \bm{1}\), and further restrict each cluster to be \emph{equally partitioned} \cite{Asano:2020ww}, which can be expressed as
\[\bm{C}^\top \bm{1} = \frac{n}{k}\bm{1}.\]
Please kindly note that here the equipartition requirement should be regarded as a regularization rather than a constraint, which aims to avoid the downgraded trivial solution. Therefore, it does not require the natural data to be in equally-sized clusters. Moreover, in experiments, we set the number of classes to be relatively larger to the real numbers (aka., the \emph{overclustering} strategy) as introduced in \citet{Caron:2018ba} and \citet{Asano:2020ww}, considering we are agnostic to the number of classes in an unsupervised setting. The overclustering strategy decomposes each data cluster into smaller sub-clusters, and thereby allows these smaller sub-clusters to be in a similar size.

\paragraph{Self-supervision on cluster assignments.}
To predict cluster labels, we employ a multilayer perception (MLP) network as the classifier. The MLP takes node embeddings \(\bm{H}\) as input and predicts correct labels on top of these embeddings. For a typical classification problem with deterministic labels, we solve the following optimization problem
\begin{equation}
	\min -\frac{1}{n} \sum_{i = 1}^n \log P(y = y_i \mid v_i),
	\label{prob:classification}
\end{equation}
where \(p(y \mid v_i) = \operatorname{softmax} (\operatorname{MLP}(\bm{h}_i))\) is the prediction for node \(v_i\). Considering that cluster assignments are relaxed to be probability distributions, Problem (\ref{prob:classification}) can be implemented as the cross-entropy loss between two distributions \(q(y \mid v_i)\) and \(p(y \mid v_i)\) \cite{Asano:2020ww}:
%which results in the following optimization problem:
%\begin{equation}
%	\min \enspace -\frac{1}{n} \sum_{i = 1}^n \log(y_i \mid v_i),
%\end{equation}

% \(h: \mathbb R^d \rightarrow \mathbb R^k\), mapping generated embeddings to clusters. It is a MLP with \(s\) layers. Let \(\bm H^{(l+k)}\) be the output of the \(k\)-th layer in \(h\).
%The layer-wise update formula is given by
%where \(\hat{\bm{C}} = \operatorname{softmax}(\operatorname{MLP}( \bm H^{(l)} ))\) is the predicted labels, \(\hat{\bm{C}} \in \mathbb R^{n \times k}\), \(k\) is the number of clusters.
%And denote the predicted labels as \(\hat{\bm C} \in \mathbb R^{n \times k}\), where \(\hat{\bm c}_i = \mathrm {softmax}(\bm h^{(l + s)}_i)\).
%We denote \(q(y\mid v_i)\) as the target distribution, which is the cluster assignments of each node, i.e. \(q(y \mid v_i) = c_{iy}\), and \(p(y \mid v_i)\) as the predicted distribution, which is the prediction of classifier, i.e. \(p(y \mid v_i) = \hat c_{iy}\). The problem in Eq. (x) can be rewritten to minimize the cross-entropy loss between two distributions \(q\) and \(p\):
\begin{equation}
	\mathcal L = - \frac{1}{n} \sum_{i=1}^n \sum_{y=1}^k q(y \mid v_i) \log p(y \mid v_i),
	\label{eq:loss}
\end{equation}
where \(q(y \mid v_i) = c_{iy}\) is the cluster assignments. Note that when \(q(y \mid v_i)\) is deterministic, minimizing Eq. (\ref{eq:loss}) is equivalent to solving Problem (\ref{prob:classification}).
%{\color{purple} new section model training? Note that the cluster assignments \(\bm C\) is fixed when updating the model parameters.} 

%\textbf{Update cluster assignments.}
%the cluster assignments of nodes are `fuzzy'. In other words, the cluster assignment of each node is a probability distribution. Assume that the cluster label set is denoted \(\mathcal L = \{ 1, 2, \cdots, k \}\), where \(k\) is the cluster number. And let \(l(i)\) denote the random variable of the cluster label of node \(v_i \in \mathcal V\). The cluster assignments \(\bm C \in \mathbb R^{n \times k}\) can be defined as a matrix with entries
%\begin{equation}
%  c_{iy} = \Pr \{ l(i) = y \}.
%\end{equation}
%Now consider the problem of cluster label update. 
%With this formulation, we minimize the cross-entropy between \(q\) and \(p\) so as to update the cluster assignments \(\bm{C}\), i.e. update \(q(y \mid v_i)\) in Eq. (\ref{eq:loss}).
With this formulation, given the current cluster assignments, we update the model parameters by minimizing the cross-entropy between \(q\) and \(p\). When updating cluster assignments, we optimize for assignments that minimize Eq. (\ref{eq:loss}) based on the currently predicted distribution \(p\).
%And to tackle the problem of \emph{degenerated solution}\cite{Caron:2018ba}, where the model assign most of the nodes into one cluster, we restrict the cluster assignment to be \emph{equipartitioned}. In other words, we place the restriction on \(\bm C\) that \(\bm C^\top \bm 1 = \frac n k\). Here, \(\bm 1\) denotes a vector with all entries being \(1\) of proper length.
According to \citet{Asano:2020ww}, this shall be formulated as an optimal transportation problem, where \(\bm{P} \in \mathbb{R}^{n \times k}\) with entries \(p_{iy} = - \log (p(y \mid v_i))\) is the cost matrix. We relax the matrix \(\bm{C}\) to be an element of the transportation polytope, given by
\begin{equation}
	U(\bm{r}, \bm{c}) := \{ \bm C \in \mathbb R_{+}^{n \times k} \mid \bm{C1} = \bm{r}, \bm{C}^\top \bm{1} = \bm{c} \},
\end{equation}
where \(\bm{r} = \bm{1} \in \mathbb{R}^n\) and \(\bm{c} = \frac{n}{k} \bm{1} \in \mathbb{R}^k\), which corresponds to our restriction of equipartition, and the solution should minimize the cost
\begin{equation}
  \langle \bm C, \bm P \rangle.
\end{equation}

This problem can be solved in near-linear time using the Sinkhorn-Knopp matrix scaling algorithm \cite{Cuturi:2013wo}. Specifically, we can solve the problem by approximating the Sinkhorn projection of \(e^{\mu \bm P}\) using the scaling algorithm, where \(\mu\) is a hyper-parameter. In CAGNN, we employ a greedy version of the Sinkhorn algorithm, Greenkhorn \cite{Altschuler:2017wa}, to approximate the solution, which is proven to outperform the original version significantly in practice. The cluster assignment updating algorithm is given in Algorithm \ref{algo:label-update}.
For the details of the Greenkhorn algorithm, we refer the readers of interest to Appendix \ref{appendix:Greenkhorn-detail}.

\begin{algorithm}[h]
	\DontPrintSemicolon
	\caption{CAGNN cluster assignment updating}
	\label{algo:label-update}
	\(\bm P \gets -\log \left(\operatorname{softmax}\left(\operatorname{MLP}\left(\bm H\right)\right)\right)\) \;
    \(\bm M \gets e^{-\eta \bm P}\)\;
    \(\bm r \gets \bm 1\)\;
    \(\bm c \gets \frac n k \bm 1\)\;
    \(\bm M \gets \texttt{Greenkhorn}(\bm M, \bm r, \bm c)\)\;
	\(\bm C \gets \bm M\)\;
\end{algorithm}

\subsection{Cluster-Aware Topology Refining}

After obtaining cluster assignments, we further refine the graph topology by strengthening intra-class edges and reducing inter-class connections. Specifically, given cluster assignments \(\bm{C}\), for each edge \((v_i, v_j)\), we remove it if the probability that \(v_i\) and \(v_j\) fall into the same cluster is less than a threshold \(\tau_r\), i.e. \(\bm c_i^\top \bm c_j < \tau_r\). Additionally, For each node pair \((v_i, v_j)\), if the probability that \(v_i\) and \(v_j\) belong to the same cluster is greater than another threshold \(\tau_a\), we add the edge \((v_i, v_j)\) to the graph.

Note that in each iteration, we refine graph topology based on the original graph instead of the previously refined graph, since informative edges might be accidentally removed at the early stage of the training procedure. Additionally, when adding edges, we consider \((v_i, v_j)\) as candidates only when \(\operatorname{argmax}_{k} c_{ik} = \operatorname{argmax}_{l} c_{jl}\) to reduce computational complexity and promote parallel computation.

The topology refining procedure is designed to increase the \emph{purity} of the whole graph \(\phi_{\mathcal{G}}\), which is defined as the probability of an edge in \(\mathcal{G}\) connecting nodes from the same cluster. Formally, we define graph purity as
\begin{equation}
	\phi_p(\mathcal G) = \frac 1 {|\mathcal E|} \sum_{(v_i, v_j) \in \mathcal E} P(y_i = y_j) = \frac 1 {|\mathcal E|} \sum_{(v_i, v_j) \in \mathcal E} \bm c_i^\top \bm c_j.
\end{equation}

Graph purity depicts how likely an edge in the graph is to connect nodes from the same cluster. We can see that topology refining can increase graph purity when the threshold is less than or equal to the current purity, i.e. \(\tau \le \phi_{p}(\mathcal G)\). In practice, \(\tau\) is chosen dynamically based on current cluster assignments with \(\tau = \frac 1 2 \phi_{p}(\mathcal G)\). 

\begin{figure}
	\centering
	\subfloat[Original]{
		\includegraphics[width=0.3\textwidth]{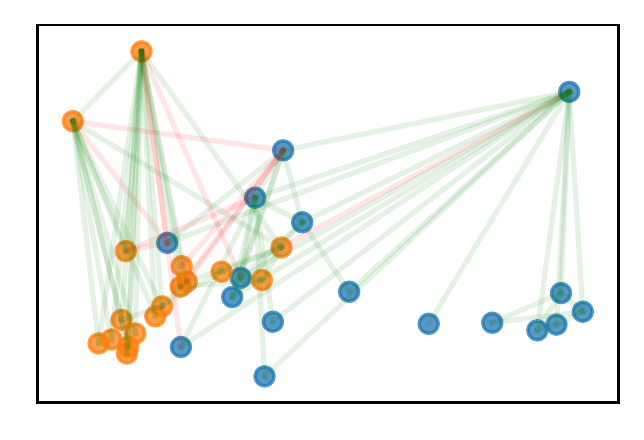}
		\label{fig:original-Karate}
	}
	\enspace
	\subfloat[Purity increased]{
		\includegraphics[width=0.3\textwidth]{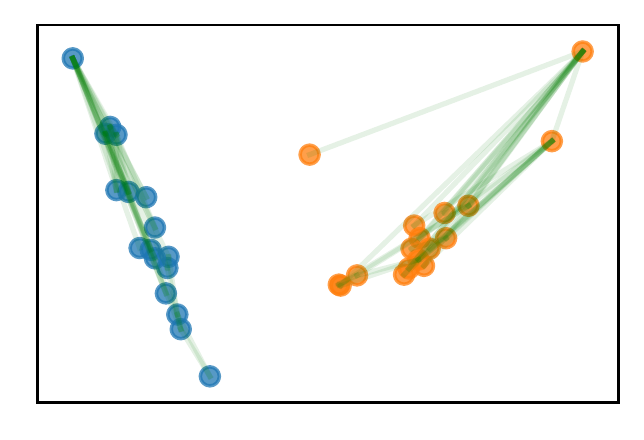}
		\label{fig:purity-increased}
	}
	\enspace
	\subfloat[Purity decreased]{
		\includegraphics[width=0.3\textwidth]{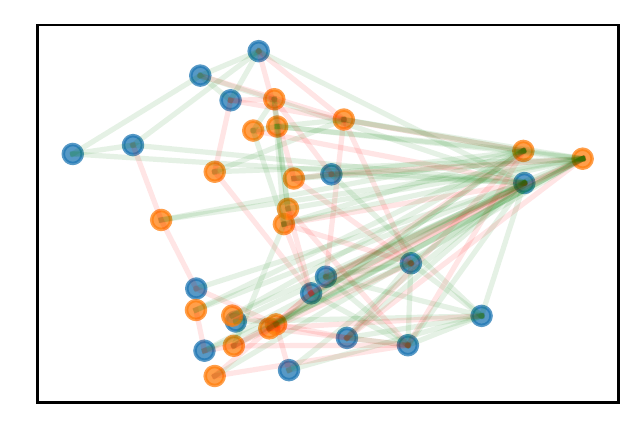}
		\label{fig:purity-decresed}
	}
	\caption{Visualization of node embeddings with different graph purity on the Karate club dataset. Node colors indicate classes. Green lines indicate inter-class edges while red lines indicate intra-class edges.}
	\label{fig:visualization}
\end{figure}

Our motivation is that graph with higher purity will generate embeddings that better preserve cluster structures. Considering that embeddings of neighboring nodes are smoothed in graph convolutions, embeddings of nodes belonging to different clusters will become similar due to inter-cluster edges, resulting in indistinctive node embeddings. We further illustrate this idea through visualization. On the Karate club dataset, we conduct cluster-aware topology refining to increase graph purity and decrease graph purity by adding noise edges. The learned embeddings are shown in Fig. \ref{fig:visualization}. We can see that the modified graph (\ref{fig:purity-increased}) with higher purity produces embeddings with well-separated clusters, while the clusters are indistinguishable in the graph (\ref{fig:purity-decresed}) with lower purity.

\subsection{Model Training and Complexity Analysis}

To train the proposed CAGNN model, we first initialize the parameters of GCN by training it with reconstruction loss, which forces the node embeddings to preserve pairwise similarity. Then, we initialize the cluster assignments by running \(k\)-Means on the node embeddings. After initialization, the node embeddings will be improved in further steps using self-supervised techniques. Specifically, we iteratively update model weights in three stages, i.e. graph representation learning, cluster assignment updating, and cluster-aware topology refining.

Note that when we perform graph representation learning by minimizing Eq. (\ref{eq:loss}), we maintain the cluster assignments \(\bm{C}\) fixed. Each time cluster assignments are updated, we perform neighborhood refining based on the new assignments. Following \citet{Asano:2020ww}, to avoid unstable self-supervised representation learning, we distribute label assignment updating throughout the whole training process. We denote \(\mathcal{S}\) as the set of epochs where cluster assignments will be updated. \(\mathcal{S}\) can be chosen freely as long as label updating is performed at proper intervals. In our implementation, we set updating epoch \(s_i \in \mathcal{S}\) as \(s_i = (E - W) \frac{i}{U + 1} + W, \ i = 1, 2, \dots, U \), where \(U\) is the total number of updates throughout training, \(W\) is the number of warm-up epochs where label updating will not be performed, and \(E\) is the total number of training epochs. The training algorithm is summarized in Algorithm \ref{algo:CAGNN-training}.

\begin{algorithm}[h]
	\DontPrintSemicolon\SetNoFillComment
	\caption{CAGNN training algorithm}
	\label{algo:CAGNN-training}
	\(\mathcal E_0 \gets \mathcal E\)\;
	Initialize the weights of GCN\;
	Generate initial embedding \(\bm{H}\) using GCN with Eq. (\ref{eq:gcn-rule})\;
	Initialize initial cluster assignments using \(k\)Means\;
	\For {\(epoch \gets 1, 2, \dots\)} {
		Update weights of GCN and MLP with Eq. (\ref{eq:loss})\;
		\If {\(epoch \in \mathcal S\)} {
		Update cluster assignments using the Sinkhorn-Knopp algorithm\;
		\tcc{Perform topology refining}
        \(\tau_r \gets \frac 1 2 \phi_p(\mathcal G) \)\;
        \(\mathcal E \gets \{ (v_i, v_j) \mid (v_i, v_j) \in \mathcal E_0, \bm c_i^\top \bm c_j \ge \tau_r \}\)\tcp*{Remove inter-class edges}
        \For{\(y \gets 1, 2, \cdots, k\)} {
          \(\mathcal V_{y} \gets \{ v_i \in \mathcal V \mid \operatorname{argmax}_{l} c_{il} = y \}\)\;
          \(\mathcal E \gets \mathcal E \cup \{ (v_i, v_j) \in \mathcal V_y \times \mathcal V_y, \bm c_i^\top \bm c_j > \tau_a \}\)\tcp*{Add intra-class edges}
        }
        }
	}
\end{algorithm}

The time complexity of updating cluster assignments using the Sinkhorn-Knopp algorithm is \(O(nk)\). In the cluster-aware topology refining procedure, we compute the correlation between each connected node pair and delete edges with low correlation, which has the time complexity of \(O(|\mathcal{E}|(k+1))\). Note that in the real world, graphs are usually sparse, i.e. \(|\mathcal{E}| \ll n^2\). Therefore, the overall time complexity of each cluster updating iteration is \(O(nk + (k+1)|\mathcal{E}|) \).

\section{Experiments}

In this section, we present the results and analysis of empirical evaluation of our proposed method. These experiments are conducted to answer the following four research questions:
\begin{itemize}
	\item \textbf{RQ1}: How does the proposed method compare with existing baselines in traditional graph mining tasks?
	\item \textbf{RQ2}: How does the cluster-aware topology refining mechanism help improve the quality of node embeddings? How does adding intra-class edges and removing inter-class edges independently contribute to improve the quality of node embeddings?
	\item \textbf{RQ3}: Does the soft topology refining scheme outperform the proposed hard refining scheme?
	\item \textbf{RQ4}: How do key hyper-parameters affect model performance?
\end{itemize}

To answer RQ1, in the experiments, we extensively compare the proposed CAGNN for two traditional graph mining tasks, node classification and node clustering.
Then, we conduct detailed ablation studies on the cluster-aware topology refining procedure to answer RQ2.
Following the ablation study of the cluster-aware topology refining module, we further compare the proposed hard refining scheme with its ``soft'' variant to answer RQ3.
After that, to answer RQ4, we perform parameter sensitivity analysis on several key hyper-parameters of the model. Finally, we provide visualization of node embeddings to give qualitative results of our proposed methods.

\subsection{Experimental Setup}

\paragraph{Datasets.}
For a comprehensive comparison with state-of-the-art methods, we evaluate our model using three widely-used citation networks, Cora, Citeseer, and Pubmed, for predicting article subject categories, provided by \citet{Sen:2008gm,Yang:2016ts}.
In these three datasets, graphs are constructed from computer science papers of various subjects. Specifically, nodes correspond to articles and undirected edges correspond to citation links. Each node has a sparse 0/1 bag-of-words feature and a corresponding class label. The statistics is summarized in Table \ref{tab:dataset-statistics}.

%\begin{itemize}
%	\item Citation networks: we use three standard citation networks: Cora, CiteSeer and PubMed. 
%	\item Document networks: we also evaluate our model on Wiki, where nodes are documents and two nodes are connected if there are links between them. Each node has a bag-of-words feature and a class label.
%\end{itemize}

\begin{table}[t]
	\centering
	\caption{Statistics of datasets used throughout experiments.}
	\label{tab:dataset-statistics}
	\resizebox{0.56\textwidth}{!}{
	\begin{tabular}{ccccc}
	\toprule
	Dataset & \#Nodes & \#Edges & \#Features & \#Classes  \\
	\midrule
	Cora & 2,708 & 5,429 & 1,433 & 7 \\
	Citeseer & 3,327 & 4,732 & 3,703 & 6 \\
	Pubmed & 19,717 & 44,338 & 500 & 3 \\
	\bottomrule
	\end{tabular}
	}
\end{table}

\paragraph{Experimental configurations.}
We train the model using the Adam optimizer with a learning rate of \(0.01\). Initially, we train the GCN model with the reconstruction loss for 500 epochs on Cora and PubMed, and 250 epochs on CiteSeer. Following that, in the self-supervised learning phase, we train the whole model for 15, 60, and 50 epochs on Cora, Citeseer, and Pubmed, respectively. On all the datasets, the optimal transportation solver is run for a fixed number epochs \(E_{ot}\) and with the same hyper-parameter \(\mu\). Prior to training, we initialize the weight of the encoders by training it with reconstruction loss. Technically, we optimize the loss by negative sampling. It is also possible to initialize the parameters in GCN with a variant of reconstruction loss proposed in VGAE \cite{Kipf:2016ul}. In our experiments, we initialize our model with the standard reconstruction loss on Cora and Pubmed, while on Citeseer we use the variational reconstruction loss to better bootstrap self-supervised training.

\paragraph{Hyper-parameter settings.}
We set the dimension of the node embeddings to 64, the weight decay to 0.0008 in all datasets. For the number of clusters, to avoid trivial solutions \cite{Caron:2018ba}, we set the number of clusters to be around twice the number of ground-truth classes. Specifically, we set the number of clusters to 10, 11, and 5 on Cora, Citeseer, and Pubmed, respectively. For the set of epochs where we perform cluster assignment updating, \(W\) is set to 1, 8, and 2 in Cora, Citeseer, and Pubmed, respectively; \(U\) is set to 7, 7, and 6 in three datasets, respectively. Besides, for the optimal transportation solver, \(E_{ot}\) is set to 1,000 and \(\mu\) is set to 20.

\subsection{Node Clustering (RQ1)}
To demonstrate the performance of the proposed approach, we first evaluate it on an unsupervised task; we conduct node clustering algorithms on top of the learned node embeddings. In this experiment, we employ \(k\)Means as the clustering method. We run the algorithm for ten (10) times and report the averaged performance.

\paragraph{Baselines.}
For a comprehensive comparison, we compare our methods against various unsupervised methods. These methods can be grouped into three categories.

\begin{itemize}
	\item Traditional methods that only make use of input features. We run two methods \(k\)Means and spectral clustering (SC) directly on the input features, which means that no graph structures are used at all.
	\item Network embedding methods which use graph structures only.
	\begin{itemize}
		\item DeepWalk \cite{Perozzi:2014ib} is a representative random-walk-based method, which generates node embeddings by sampling random walks on graphs and feeds them into language models.
		\item DNGR \cite{Cao:2016wp} adopts a random surfing model to capture the graph structures. These methods only utilize graph structural information and neglect the input features.
	\end{itemize}
	\item Attributed graph clustering models that use both structures and attributes.
	\begin{itemize}
		\item Graph autoencoders (GAE \cite{Kipf:2016ul}, VGAE \cite{Kipf:2016ul}, and MGAE \cite{Wang:2017go}) use GCN \cite{Kipf:2016tc} as the encoder and enforce the model to reconstruct graph structures specified by a graph proximity matrix (e.g., the adjacency matrix which represents one-order proximities).
		\item DANE \cite{Gao:2018jn} employs two autoencoders to preserve proximities for both graph structures and node attributes.
		\item AGC \cite{Zhang:2019bi} directly applies graph convolutions to the input features and runs spectral clustering on the obtained embeddings.
		\item DGI \cite{Velickovic:2019tu} applies contrastive learning techniques that aims to maximize mutual information between global graph embeddings and local node embeddings.
	\end{itemize}
\end{itemize}

\paragraph{Evaluation metrics.}
We report the performance in terms of three widely-used metrics for evaluating cluster quality, i.e. micro-averaged F1-score (Mi-F1), macro-averaged F1-score (Ma-F1), and normalized mutual information (NMI). The two F1-scores are calculated respectively as 
\begin{align}
	\text{Micro F1-score} & = 2 \times \frac{\text{Micro-Precision} + \text{Micro-Recall}}{\text{Micro-Precision} \times \text{Micro-Recall}}, \\
	\text{Macro F1-score} & = 2 \times \frac{\text{Macro-Precision} + \text{Macro-Recall}}{\text{Macro-Precision} \times \text{Macro-Recall}},
\end{align}
where Precision is the number of true positives divided by the sum of true positives and false positives, whereas Recall is the number of true positives divided by the sum of true positives and false negatives.
Note that our all datasets are in the multi-class classification setup; the macro-averaged score computes the metric independently for each class and then report the averaged score, while the micro-averaged score is obtained by aggregating the contributions of all classes.

Besides, the NMI score is another widely-adopted metric for evaluation of clustering, which is calculated by 
\begin{equation}
	\text{NMI}(Y, C) = \frac{2 \times I(Y; C)}{H(Y) + H(C)},
\end{equation}
where \(Y\) and \(C\) are true class labels and predicted cluster labels respectively, \(I(Y; C)\) measures information between \(Y\) and \(C\), and \(H(\cdot)\) measures entropy.

\paragraph{Results and analysis.}
The performance is summarized in Table \ref{tab:node-clustering} with the highest performance highlighted in bold. We report the performance of baselines in accordance with their original papers \cite{Gao:2018jn,Zhang:2019bi}. Overall, from the table, it is seen that our proposed CAGNN surpasses other baseline methods in terms of accuracy on all three datasets. It is worth mentioning that we exceed the existing state-of-the-art model by a large margin of over \(7\%\) in terms of absolute accuracy improvements on Cora.

\begin{table}[t]
	\centering
	\caption{Performance of node clustering on three citation networks in terms of Micro-F1 (Mi-F1), Macro-F1 (Ma-F1), and normalized mutual information (NMI).}
	\label{tab:node-clustering}
    \begin{tabular}{cccccccccc}
	\toprule
	\multirow{2.5}[0]{*}{Method} & \multicolumn{3}{c}{Cora} & \multicolumn{3}{c}{Citeseer} & \multicolumn{3}{c}{Pubmed} \\
	\cmidrule(lr){2-4} \cmidrule(lr){5-7} \cmidrule(lr){8-10}
		& Mi-F1 & Ma-F1  & NMI   & Mi-F1 & Ma-F1  & NMI   & Mi-F1  & Ma-F1 & NMI   \\
	\midrule
    \(k\)-Means & 34.65 & 25.42 & 16.73 & 38.49 & 30.47 & 17.02 & 33.37 & 57.35 & 29.12 \\
    SC & 36.26 & 25.64 & 15.09 & 46.23 & 33.70 & 21.19 & 59.91 & 58.61 & 32.55 \\
    \midrule
    DeepWalk & 46.74 & 38.06 & 31.75 & 36.15 & 26.70 & 9.66  & 61.86 & 47.06 & 16.71 \\
    DNGR  & 49.24 & 37.29 & 37.29 & 32.59 & 44.19 & 18.02 & 45.35 & 17.90 & 15.38 \\
    \midrule
    GAE   & 53.25 & 41.97 & 40.69 & 41.26 & 29.13 & 18.34 & 64.08 & 49.26 & 22.97 \\
    VGAE  & 55.95 & 41.50 & 38.45 & 44.38 & 31.88 & 22.71 & 65.48 & 50.95 & 25.09 \\
    MGAE  & 63.43 & 38.01 & 45.57 & 63.56 & 39.49 & 39.75 & 43.88 & 41.98 & 8.16 \\
    ARGE  & 64.00 & 61.90 & 44.90 & 57.30 & 54.60 & 35.00 & 59.12 & 58.41 & 23.17 \\
    ARVGE & 63.80 & 62.70 & 45.00 & 54.40 & 52.90 & 26.10 & 58.22 & 23.04 & 20.62 \\
    DANE  & 70.27 & 68.93 & 55.15 & 47.97 & 45.28 & 24.25 & 69.42 & 65.10 & 29.30 \\
    AGC   & 68.92 & 65.61 & 53.68 & 67.00 & \textbf{62.48} & \textbf{41.13} & 69.78 & 68.72 & 31.59 \\
    DGI   & 65.28 & 58.90 & 47.85 & 60.37 & 55.63 & 38.81 & 51.22 & 46.73 & 18.54 \\
    \midrule
    \textbf{CAGNN} & \textbf{77.37} & \textbf{75.24} & \textbf{58.90} & \textbf{67.30} & 62.20 & 40.84 & \textbf{71.03} & \textbf{70.72} & \textbf{36.09} \\
	\bottomrule
 	\end{tabular}
\end{table}

The results can be analyzed in three aspects. First of all, we observe that traditional algorithms such as \(k\)Means and spectral clustering, which simply rely on node attributes perform poorly on graph data.
Secondly, conventional network embedding methods outperform traditional clustering methods, but their performance is still inferior to attributed graph clustering methods. This demonstrates the power of modern deep learning techniques on graphs, that can better leverage both graph structures and node attributes.
Last, it is observed that our proposed method outperforms attributed graph clustering baselines by considerable margins. Previous graph clustering methods merely perform node representation learning on the node level, while our method exploits underlying cluster structures in graph to guide representation learning. Additionally, we utilize cluster information to reduce the impact of noisy inter-class edges, which further benefits the quality of embeddings. The improvements show that our proposed cluster-aware self-supervision learning method and topology refining scheme help generate embeddings that better preserve cluster structures.
%while the proposed method further considers guiding the GCN training by reducing the impact of noisy inter-class edges. The improvements show that the proposed topology refining scheme is able to better preserve cluster structures in the embedding space.

Note that the proposed CAGNN is slightly inferior to AGC on Citeseer in terms of NMI and Macro-F1 score. However, CAGNN still outperforms it in terms of accuracy and on other datasets by a considerable margin. In all, these results verify the effectiveness of our proposed CAGNN.

\subsection{Node Classification (RQ1)}
We further evaluate the quality of embeddings generated by CAGNN on a supervised task, i.e. node classification. After training the model, we conduct node classification on the learned node representations. For a fair comparison, we closely follow the same experimental settings as \cite{Gao:2018jn}. Specifically, we train a linear logistic regression classifier with \(\ell_2\) regularization. For training/test set splitting, we randomly select 10\% nodes as the training set and the remaining nodes are left for the test set. Then, we use five-fold cross-validation to select the best model. We report the performance on the test set in terms of two widely-used metrics, Micro-averaged F1-score (Mi-F1) and Macro-averaged F1-score (Ma-F1). As with the previous experiment, we report the averaged performance of ten (10) runs.

\paragraph{Baselines.}
In node classification, we include two lines of baseline algorithms: (1) traditional network embedding methods, which only leverage graph structures and ignore node attributes, and (2) attributed graph representation learning methods, which use both graph structures and node attributes.
The former category includes representative random-walk-based methods DeepWalk \cite{Perozzi:2014ib}, node2vec \cite{Grover:2016ex}, and GraRep \cite{Cao:2015ie}. The latter one includes attributed network embedding methods ANE \cite{Huang:2017uy} and DANE \cite{Gao:2018jn}, and graph neural networks GAE, VGAE \cite{Kipf:2016ul}, and DGI \cite{Velickovic:2019tu}.

\paragraph{Results and analysis.}
We report the performance in Table \ref{tab:node-clf}, with the best performance highlighted in boldface. The baseline performance is reported as in their original papers \cite{Gao:2018jn}. In general, the results confirm the effectiveness of the proposed method. As shown in the table, the proposed CAGNN performs best on the Cora and Citeseer datasets, compared with state-of-the-art baselines, and shows competitive performance on the Pubmed compared with other graph representation learning methods.

\begin{table}[b]
	\centering
	\caption{Performance of node classification on three citation networks in terms of Micro-F1 (Mi-F1) and Macro-F1 (Ma-F1).}
    \begin{tabular}{ccccccc}
	\toprule
	\multirow{2.5}[0]{*}{Method} & \multicolumn{2}{c}{Cora} & \multicolumn{2}{c}{Citeseer} & \multicolumn{2}{c}{Pubmed} \\
	\cmidrule(lr){2-3} \cmidrule(lr){4-5} \cmidrule(lr){6-7}
		& Mi-F1 & Ma-F1 & Mi-F1 & Ma-F1 & Mi-F1 &  Ma-F1 \\
	\midrule
    DeepWalk & 75.68 & 74.98 & 50.52 & 46.45 & 80.47 & 78.73\\
    node2vec & 74.77 & 72.56 & 52.33 & 48.32 & 80.27 & 78.49\\
    GraRep & 75.68 & 74.41 & 48.17 & 45.89 & 79.51 & 77.85\\
    \midrule
    ANE & 72.03 & 71.50 & 58.77 & 54.51 & 79.77 & 78.75\\
    GAE & 76.91 & 75.73 & 60.58 & 55.32 & 82.85 & 83.28\\
    VGAE & 78.88 & 77.36 & 61.15 & 56.62 & 82.99 & 82.40\\
    DANE & 78.67 & 77.48 & 64.44 & 60.43 & \textbf{86.08} & \textbf{85.79}\\
    DGI  & 82.53 & 81.09 & 68.76 & \textbf{63.58} & 85.98 & 85.66 \\
    \midrule
    CAGNN & \textbf{82.56} & \textbf{81.16} & \textbf{69.56} & 61.59 & 85.76 & 83.49 \\
	\bottomrule
 	\end{tabular}
	\label{tab:node-clf}
\end{table}

As with the conclusions drawn from the experiment of node clustering, traditional network embedding methods such as DeepWalk and node2vec perform worse than deep-learning-based graph representation learning methods, which highlights the importance of incorporating node attributes when training the model. Instead of merely leveraging graph structures, graph neural networks combine information of graph topology and node attributive information, resulting in better node embeddings.
Our proposed method further utilizes cluster information in representation learning and refines graph topology by removing inter-class edges that potentially hinder the model from preserving cluster structures in the embedding space. The proposed method produces better node embeddings, so that it achieves significant improvement over existing GCN-based methods.

Note that while DANE is a strong baseline on Pubmed, our proposed CAGNN still outperforms it in terms of accuracy and Macro-F1 score on the other two datasets.
%This may be explained by the fact that the scale of Pubmed is much larger than that of Citeseer. Through the topology refining procedure, our proposed CAGNN may accidentally remove informative edges, which results in performance loss.
We observe that the NMI between cluster labels and ground-truth labels on Pubmed is the lowest among the three datasets, which can help explain the slightly inferior performance of CAGNN on Pubmed. Through the topology refining procedure that is based on cluster labels, our proposed CAGNN may accidentally remove informative edges, which results in performance loss.
Recent work DGI marries the power of contrastive learning into graph representation learning. However, it only optimizes node representations in the latent space, which neglect fine-grained cluster structures. Therefore, on graph clustering tasks, our proposed CAGNN significantly outperforms DGI, and achieves comparable performance in node classification.

\subsection{Ablation Studies (RQ2)}

To further validate the proposed cluster-aware topology refining procedure and justify our architectural design choice, we conduct ablation studies by removing specific components in the topology refining module.
Then, we conduct node clustering using the same setting described in previous sections.

\subsubsection{Impact of the proposed topology refining module.}

\begin{figure}[t]
	\centering
	\includegraphics[width=0.75\textwidth]{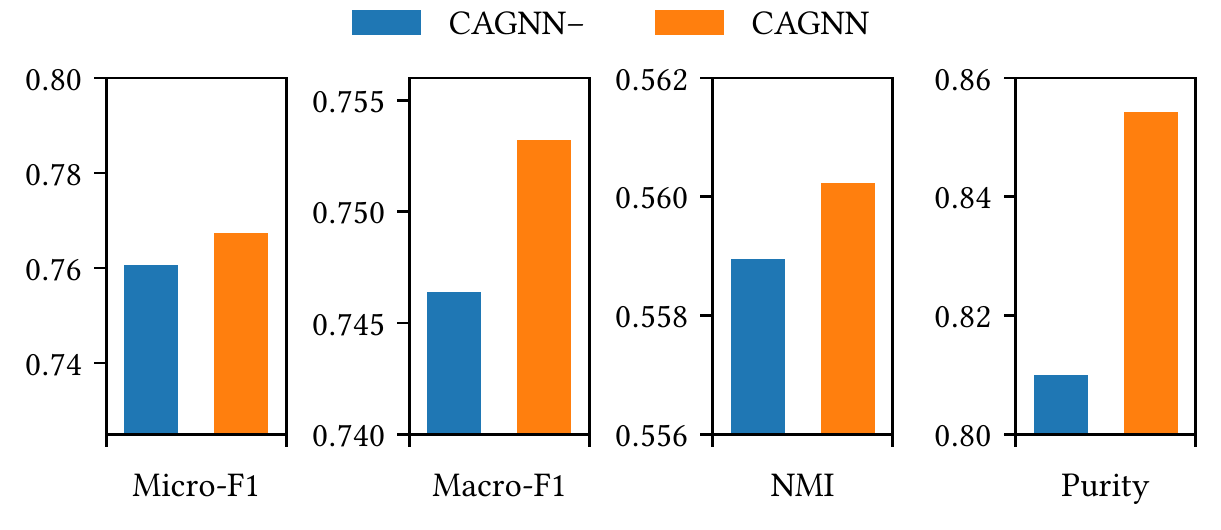}
	\caption{Performance of node clustering and graph purity on the Cora dataset with and without the topology refining module.}
	\label{fig:ablation}
\end{figure}

Firstly, to further validate the proposed cluster-aware topology refining module, we conduct ablation studies by removing this module. We term the resulting model as CAGNN-- hereafter.
To compare the performance of the original CAGNN and CAGNN--, we conduct node clustering using the same setting described in previous sections, where the performance is reported in Fig. \ref{fig:ablation}. From the figure, we observe that the topology refining module improves the performance of CAGNN-- on node clustering by considerable margins in terms of three evaluation metrics, i.e. Micro-F1, Macro-F1, and NMI, which once again verifies its effectiveness.
Moreover, we calculate graph purity against ground-truth classes to reflect the modification to topology of the original graph. From the figure, it is apparent that the proposed topology refining procedure is able to alleviate the impact of noisy inter-class edges and further better preserve cluster structures.

\subsubsection{Impact of schemes in the topology refining module.}

\begin{figure}[t]
	\centering
	\includegraphics[width=0.75\textwidth]{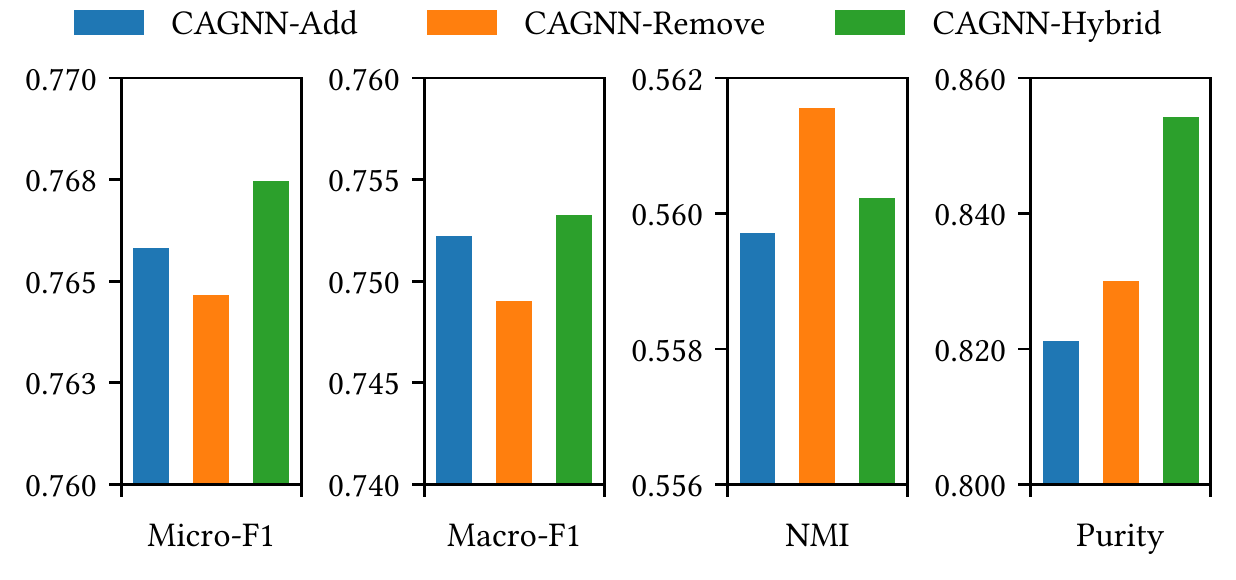} 
	\caption{Performance of node clustering on Cora with different schemes enabled in the topology refining module.}
	\label{fig:ablation-details}
\end{figure}

To further validate the proposed topology refining schemes, we perform ablation studies by comparing the model performance with different components of the refining module enabled. We report the clustering accuracy of the following three variants: (1) CAGNN-Add, which only adds intra-class edges, (2) CAGNN-Remove, which only removes inter-class connections, and (3) CAGNN-Hybrid, which is our proposed module with both schemes enabled. The performance of the three variants is presented in Fig. \ref{fig:ablation-details}.

From the figure, it is clear that enabling both schemes benefits model performance in terms of Micro-F1, Macro-F1, and Purity. However, we note that, CAGNN-Remove outperforms CAGNN-Hybrid in terms of NMI slightly. This may be explained from the fact that CAGNN-Hybrid introduces some noisy edges when adding intra-class edges, as the model makes wrong prediction about the ground-truth classes. In summary, the proposed hybrid scheme generally outperforms better, compared with CAGNN-Add and CAGNN-Remove, which justifies our design choice of the proposed topology refining module.

\begin{figure}[t]
	\centering
	\includegraphics[width=0.75\textwidth]{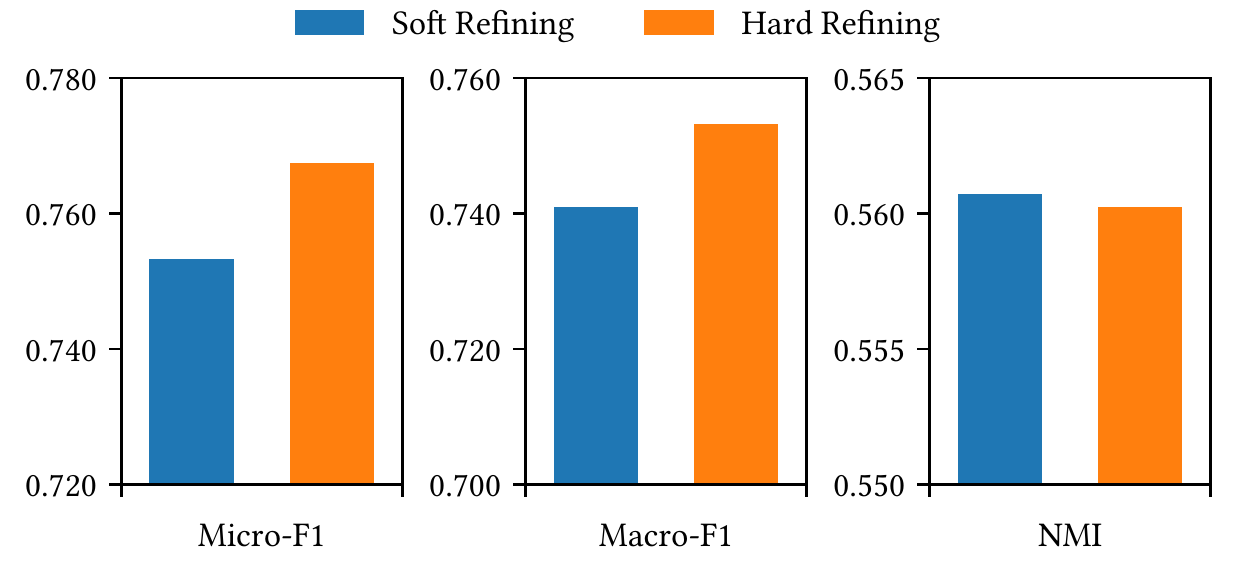} 
	\caption{Performance of node clustering on Cora with hard and soft topology refining schemes.}
	\label{fig:hard-soft}
\end{figure}

\subsection{Discussions of Hard and Soft Topology Refining Schemes (RQ3)}
Following the ablation study of the proposed topology refining scheme, we further conduct additional experiments using a soft topology refining scheme to answer RQ3.
For the proposed topology refining scheme, we regard the edge deletion as a ``hard'' operation, where the intra-class edges will be \emph{completely} removed for node representation learning. Considering the discrepancy between our model prediction about clusters and ground-truth labels, contrary to hard removal, we may consider an alternative ``soft'' scheme, where one intra-class edge are reassigned probabilities that express the strength of connection. In this experiment, for each edge \((v_i, v_j)\), we reassign each intra-class edge with a weight \(\bm{A}'_{ij} = \bm{c}_i^\top \bm{c}_j\); other edge weights are not modified. Since in the original graph, we represent each edge by \(\bm{A}_{ij} = 1\), our soft modification \(\bm{A}'_{ij} < 1\) which is able to reduce the connection of intra-class nodes.

The results are shown in Fig. \ref{fig:hard-soft}.
We empirically observe that our proposed hard scheme evidently outperforms its soft variant in terms of Micro-F1 and Macro-F1, and performs slightly lower in terms of NMI.
The result provides the rationale of using a hard removal scheme.
The reason why the soft topology refining scheme performs worse than the hard scheme may be explained from that via the soft removal scheme, there are still many inter-class edges remained, which deteriorate the quality of node embeddings.

\subsection{Parameter Sensitivity Analysis (RQ4)}
In this section, we examine the impact of two key parameters in our model, i.e. the cluster size, the two thresholds for topology refining, and the time interval of updating the clusters. We conduct node clustering on the Cora dataset by varying these four parameters independently. While one hyper-parameter studied in the sensitivity analysis is changed, other hyper-parameters remain the same as previously described.

\begin{figure}[t]
	\centering
	\subfloat[Varied numbers of clusters]{
		\includegraphics[width=0.4\textwidth]{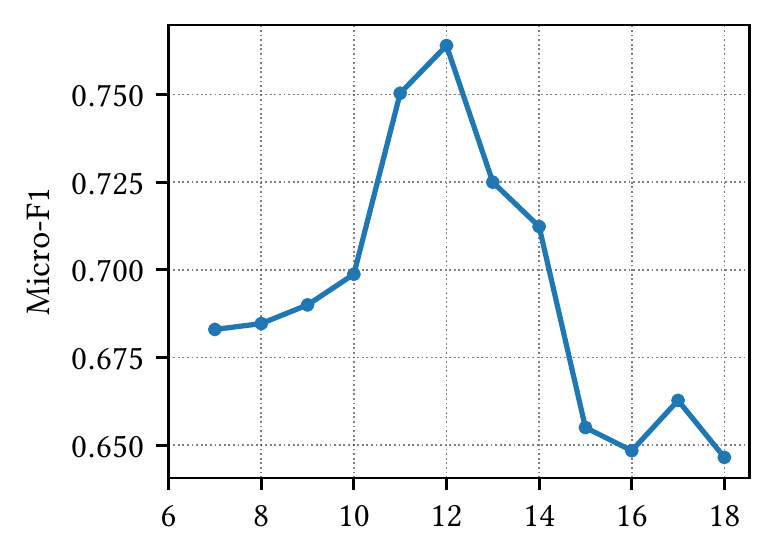}
		\label{fig:sensitivity-cluster}
	}
	\quad
	\subfloat[Varied thresholds \(\tau\)]{
		\includegraphics[width=0.4\textwidth]{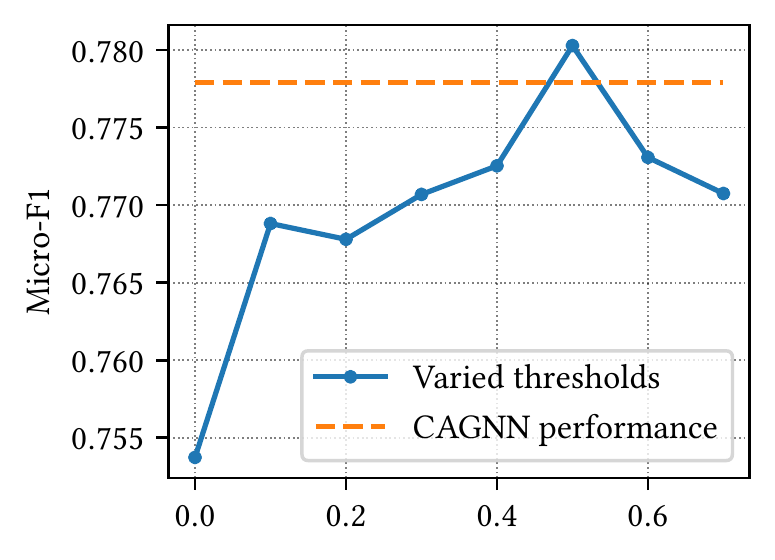}
		\label{fig:sensitivity-tau}
	}\\
	\subfloat[Varied thresholds \(\tau_a\)]{
		\includegraphics[width=0.4\textwidth]{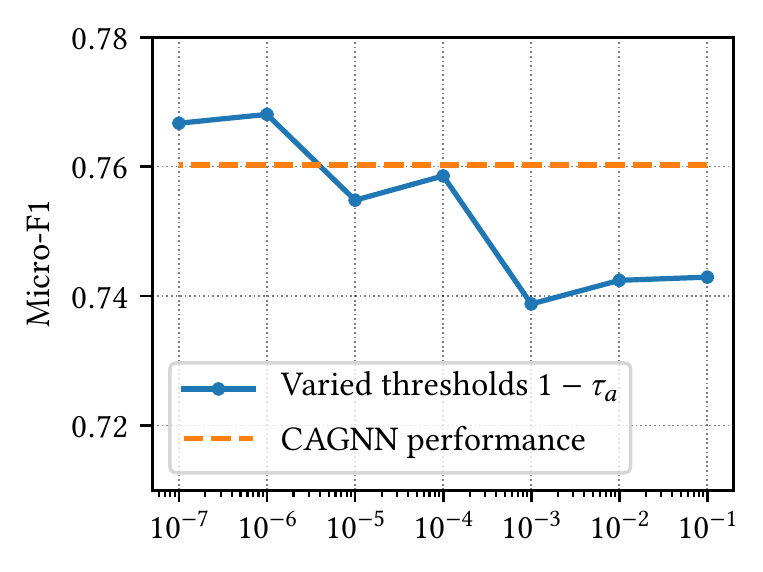}
		\label{fig:sensitivity-eta}
	}
	\quad
	\subfloat[Varied intervals of updating clusters]{
		\includegraphics[width=0.4\textwidth]{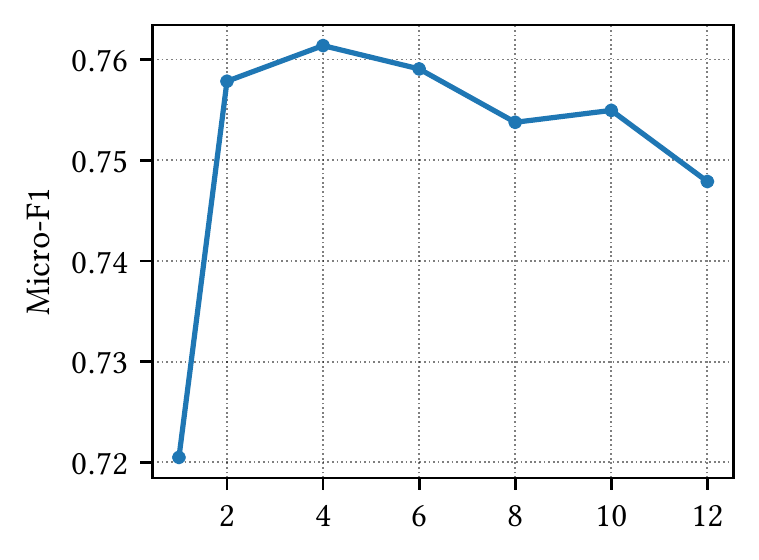}
		\label{fig:sensitivity-reassign}
	}
	\caption{Sensitivity analysis under different cluster sizes, thresholds \(\tau\), \(\tau_a\), and cluster updating intervals \(U\), in terms of node clustering accuracy on the Cora dataset.}
	\label{fig:sensitivity}
\end{figure}

\subsubsection{Impact of the cluster size.}
To investigate the influence of cluster numbers on our model, we run our model by varying the number of clusters from 7 to 18. The results on Cora with different numbers of clusters are plotted in Fig. \ref{fig:sensitivity-cluster}. From the figure, we observe that the model performance first benefits from the increase of cluster numbers, but soon the performance decreases. This indicates that the over-clustering strategy does boost the performance of UGNN, since it can alleviate inconsistency between our self-supervised learning scheme and real-world datasets. Specifically, when we enforce each cluster to be equally balanced, classes in real-world graphs usually vary greatly in their sizes. However, dividing nodes into too many clusters will in turn deteriorate the performance, since the proposed cluster-aware topology refining mechanism will unnecessarily remove informative inter-cluster edges.

\subsubsection{Impact of the threshold \(\tau_r\) in topology refining.}
To further investigate the impact of \(\tau\) on the model performance, we run CAGNN by setting \(\tau\) from 0 to 0.7, with a constant interval of 0.1. From the results in Fig. \ref{fig:sensitivity-tau}, we observe that clustering accuracy is first boosted from the increase of \(\tau\), then it stops increasing and decreases. This can be explained that a higher threshold may result in the accidental removal of possibly useful intra-cluster edges. The observation is consistent with our previous study on the impact of cluster size. 
%Moreover, we note that the highest performance when varying the threshold is obtained when the threshold is close to the ground-truth graph purity. Since the performance of the proposed CAGNN is comparable to the highest performance, the results demonstrate the effectiveness of our dynamical selection strategy by setting the threshold according to current graph topology, since in an unsupervised setting, we are not able to calculate the ground-truth graph purity.
Moreover, we note that the performance achieved with our proposed scheme that selecting \(\tau\) dynamically is close to the highest performance when directly fixing \(\tau\) to a certain value, which prove the validity of the dynamic selection scheme. Since in the real world, ground-truth labels may be inaccessible, it is infeasible to fix \(\tau\) to be the best value based on performance.

\subsubsection{Impact of the threshold \(\tau_a\) in topology refining.}
To investigate the impact of \(\tau_a\) on the performance of CAGNN, we run CAGNN by setting \(\tau_a\) to different values and report the clustering accuracy on Cora. Due to the high sparse nature of edges in the original graph, we set \(1 - \tau_a\) from \(10^{-1}\) to \(10^{-7}\) by taking exponential scales. We report  the performance under different \(\tau_a\) in Fig. \ref{fig:sensitivity-eta}.
From the figure we can see that, model performance first benefits from the increase of \(1 - \tau_a\) (which means adding more edges), but soon the accuracy decreases. The performance gain when \(1 - \tau_a\) is set to \(10^{-7}\) or \(10^{-6}\) verifies the effectiveness of our proposed adding edge scheme for topology refining. While the model benefits from the adding edge scheme initially, the performance becomes inferior to the base model when \(1 - \tau_a\) is large. This can be attributed to the fact that a large \(1 - \tau_a\) will result in a dense neighborhood, which leads to the over-smoothing problem and tends to bring noise into node representations.

\subsubsection{Impact of the cluster assignment updating interval \(U\).}
To investigate the impact of the interval of cluster assignment updating \(U\), we run CAGNN by setting \(U\) to different values. The node clustering accuracies on Cora with different intervals of assignment updating are reported in Fig. \ref{fig:sensitivity-reassign}. From the results, we can make observations such that model performance first benefits from the increase of \(U\), but soon the accuracy levels off. The performance gain when \(U\) increases can be explained by the fact that assignments updating can result in more reliable pseudo-labels, which can better guide the learning of our model. This is consistent with our motivation that the learning of model and the pseudo-labels can benefit from the progress of each other and jointly boost the quality of learned representations. On the other hand, updating cluster assignments too frequently may bring instability to model training and thus leads to inferior model performance.

\subsection{Visualizing Node Embeddings}

\begin{figure}[t]
	\centering
	\subfloat[Raw features]{
		\includegraphics[width=0.4\textwidth]{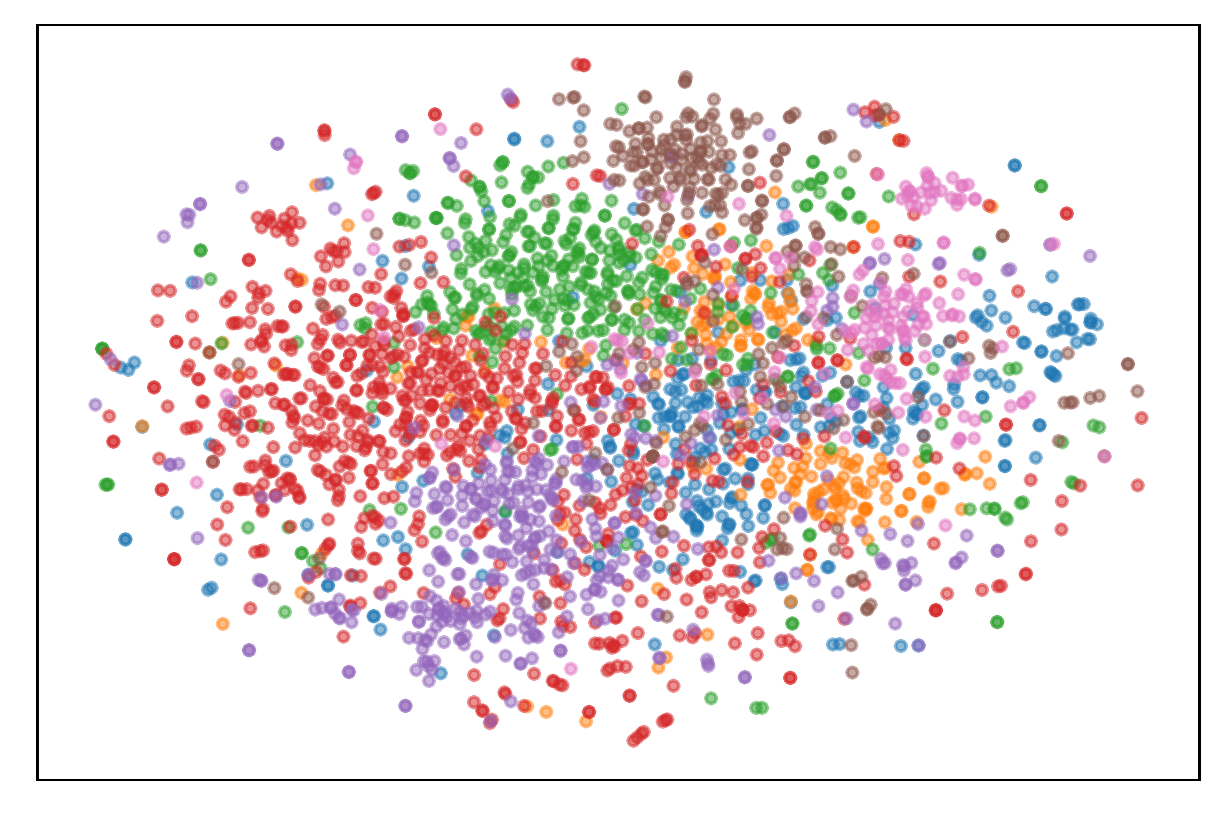}
		\label{fig:visualization-cora-raw}
	}
	\quad
	\subfloat[Node embeddings learned by CAGNN]{
		\includegraphics[width=0.4\textwidth]{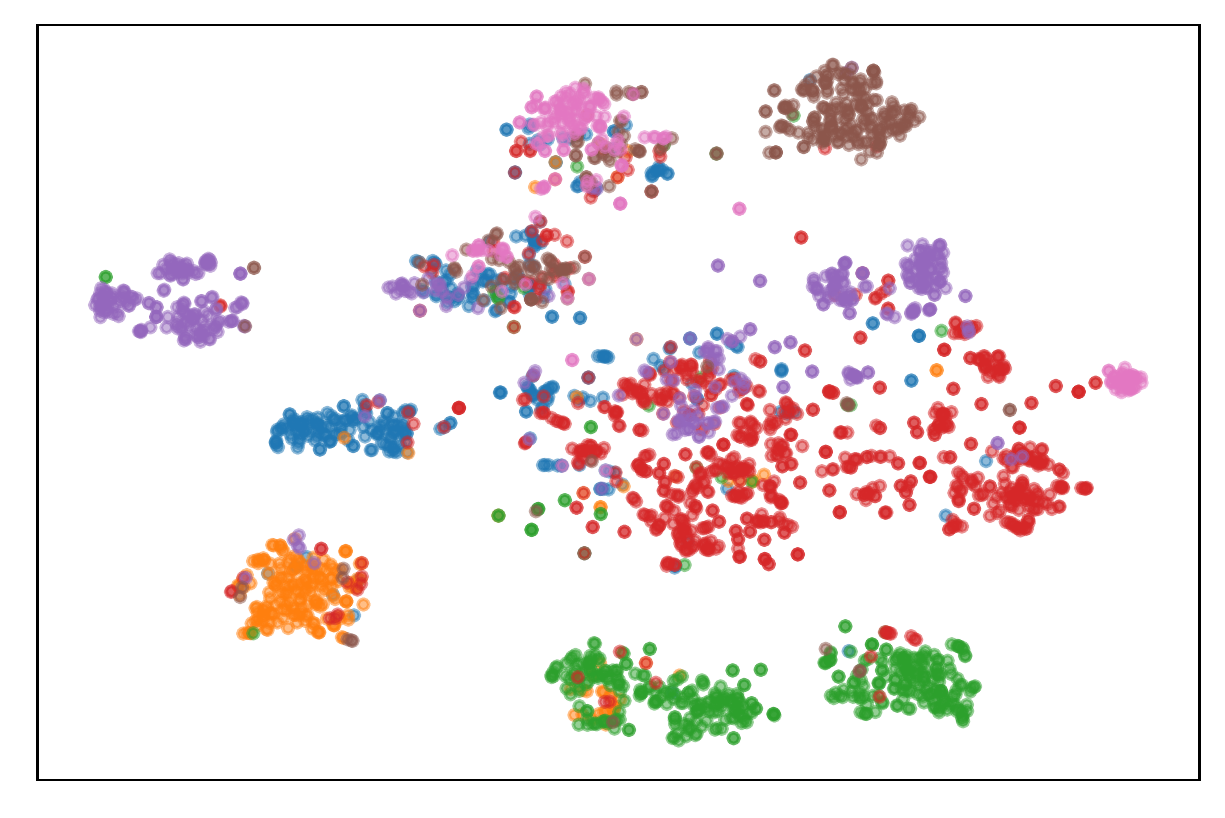}
		\label{fig:visualization-cora-cagnn}
	}
	\caption{Visualization of raw features and embeddings learned with CAGNN on the Cora dataset. T-SNE \cite{vanderMaaten:2008tm} is applied to project features and embeddings into two-dimensional spaces. Each node is colored with its corresponding ground-truth class label.}
	\label{fig:visualization-embedding}
\end{figure}

Finally, we provide qualitative results by visualizing the learnt embeddings. Specifically, we leverage t-SNE \cite{vanderMaaten:2008tm} to project the embeddings on to a two-dimensional space and plot them on a plane, colored according to the class label of each node. The node embeddings are extracted from the penultimate layer of a CAGNN model that is pre-trained on the Cora dataset.
%As is seen in Figure \ref{fig:visualization-embedding}, the embeddings learned with CAGNN exhibit much better cluster structures compared to raw features. This further verify the effectiveness of CAGNN and shows that our model can produce embeddings that well preserve the underlying cluster structures of the graph.
For comparison, we also present the visualization of raw node features of the Cora dataset in Fig. \ref{fig:visualization-cora-raw}. As is seen in Fig \ref{fig:visualization-cora-cagnn}, the representations learned with CAGNN exhibit discernible clusters in the projected two-dimensional space. Note that node colors correspond to seven \emph{ground-truth} node classes, which shows that the produced embeddings are highly discriminative across seven classes in Cora. The much clearer cluster structures of learned representations compared to raw features verify that CAGNN is able to extract useful information from graphs and preserve the underlying structure of the graph very well.

\section{Conclusion and Future Work}
In this paper, we have developed a novel cluster-aware graph neural network (CAGNN) model for unsupervised graph representation learning, in which we employ graph neural networks in a self-supervised manner. CAGNN performs clustering on the node embeddings and updates parameters by predicting cluster assignments of nodes. Moreover, we propose a novel graph topology refining scheme which strengthens intra-class edges and isolates nodes from different clusters based on cluster labels. Comprehensive experiments on two benchmark tasks using real-world datasets have been conducted. The results demonstrate the superior performance of our proposed CAGNN over state-of-the-art baselines.

The study of self-supervised techniques in graph representation learning generally remains widely open. It is seen from this work that accurately predicting the cluster labels is crucial for successfully deploying the model. In our future work, we plan to further investigate combining other self-supervised methods, e.g., contrastive learning methods, to help better model the latent space of node embeddings, and thereby improve the quality of node embeddings.

\begin{acks}
This work is jointly supported by National Key Research and Development Program (2018YFB1402600, 2016YFB1001000) and National Natural Science Foundation of China (U19B2038, 61772528).
\end{acks}

\appendix
\section{Details of the Greenkhorn Algorithm}
\label{appendix:Greenkhorn-detail}

The Greenkhorn algorithm \cite{Altschuler:2017wa} aims to solve the matrix scaling problem: given a non-negative matrix \(\bm{A} \in \mathbb{R}^{n \times k}_{+}\), the goal is to find two vectors \(\bm{x} \in \mathbb{R}^{n}, \bm{y} \in \mathbb{R}^{k}\), such that the row sum and the column sum in \(\bm{M} = \operatorname{diag}(\bm{x}) \bm{A} \operatorname{diag}(\bm{y})\) satisfy that
\begin{align}
	r(\bm{M}) & = \bm{r}, \\
	c(\bm{M}) & = \bm{c},
\end{align}
where \(r(\bm{M}) = \bm{M1}\), \(c(\bm{M}) = \bm{M}^\top\bm {1}\), \(\bm{r}\) and \(\bm{c}\) is the required row/column sum.

The vanilla Sinkhorn-Knopp algorithm approximates the solution by alternatively normalizing the row and column sum of the matrix. Instead of normalizing all rows/columns at each iteration, Greenkhorn greedily selects one row or column to update according to a distance function, \(\rho: \mathbb R^+ \times \mathbb R^+ \rightarrow \mathbb [0, +\infty]\), which is defined as
\begin{equation}
  \rho(a, b) = b - a + a \log \frac a b.
\end{equation}
The details of the Greenkhorn algorithm are given in Algorithm \ref{algo:greenkhorn}, where \(E_{ot}\) is the number of iterations.

\begin{algorithm}[h]
	\DontPrintSemicolon
	\caption{The Greenkhorn algorithm}
	\label{algo:greenkhorn}
	\SetKwFunction{FMain}{Greenkhorn}
    \SetKwProg{Fn}{function}{:}{}
    \Fn{\FMain{\(\bm A, \bm r, \bm c\)}}{
	\(\bm P \gets -\log \left(\operatorname{softmax}\left(\operatorname{MLP}\left(\bm H\right)\right)\right)\) \;
	\(\bm M^{(0)} \gets \bm A\)\;
	\( \bm x \gets \bm 0, \bm y \gets \bm 0\)\;
	\(\bm M \gets \bm M^{(0)}\) \;
	\For {epoch \(\gets\) \(1\) to \(E_{ot}\)} {
	  \(I \gets \mathrm{argmax}_{i} \  \rho(\bm r_i, r_i(\bm M))\) \;
	  \(J \gets \mathrm{argmax}_{j} \  \rho(\bm c_i, c_i(\bm M))\) \;
	  \eIf {\(\rho(\bm r_I, r_I(\bm M)) > \rho(\bm c_J, c_j(\bm M))\)} {
	  \(\bm x_I \gets \bm x_I \cdot \frac {\bm r_I} {r_I(\bm M)}\) \;
	  } {
	  \(\bm y_I \gets \bm y_I \cdot \frac {\bm c_J} {c_J(\bm M)}\) \;
	  }
	  \(\bm M \gets \mathrm{diag}(\bm x) \bm M^{(0)} \mathrm{diag}(\bm y)\) \;
	}
	\Return \(\bm M\)\;
	}
\end{algorithm}

\bibliographystyle{ACM-Reference-Format}
\bibliography{ref}

\end{document}